\documentclass{article}
\usepackage{spconf}
\usepackage{microtype}
\usepackage{amsmath,amsthm,amssymb}
\usepackage{graphicx,epstopdf,epsfig}
\usepackage{url,framed}
\usepackage{bm}
\usepackage{multirow}
\usepackage[caption=false,font=footnotesize]{subfig}
\usepackage[linesnumbered,ruled]{algorithm2e}
\usepackage{xcolor}
\usepackage{array}
\usepackage{float}
\usepackage{cite}

\def\x{\boldsymbol{x}}
\def\s{\boldsymbol{s}}
\def\n{\boldsymbol{n}}
\def\y{\boldsymbol{y}}
\def\v{\boldsymbol{v}}
\def\u{\boldsymbol{u}}

\def\g{\boldsymbol{g}}
\def\h{\boldsymbol{h}}
\def\A{\mathbf{A}}
\def\G{\mathbf{G}}
\def\K{\mathbf{K}}
\def\P{\mathbf{P}}
\def\W{\mathbf{W}}
\def\D{\mathfrak{D}}
\def\L{\mathfrak{L}}
\title{Linearized ADMM and Fast Nonlocal Denoising for Efficient Plug-and-Play Restoration}
\name{Unni~V.~S.,~Sanjay Ghosh~and~Kunal~N.~Chaudhury}
\address{Department of Electrical Engineering, Indian Institute of Science, Bengaluru, India}
\begin{document}
\ninept

% ------------------ Title ---------------------------
\maketitle
% ----------------- Abstract -------------------------
\begin{abstract}
In plug-and-play image restoration, the regularization is performed using powerful denoisers such as nonlocal means (NLM) or BM3D. This is done within the framework of alternating direction method of multipliers (ADMM), where the regularization step is formally replaced by an off-the-shelf denoiser. 
Each plug-and-play iteration involves the inversion of the forward model followed by a denoising step. In this paper, we present a couple of ideas for improving the efficiency of the inversion and denoising steps. First, we propose to use linearized ADMM, which generally allows us to perform the inversion at a lower cost than standard ADMM. Moreover, we can easily incorporate hard constraints into the optimization framework as a result. Second, we develop a fast algorithm for doubly stochastic NLM, originally proposed by Sreehari et al. (IEEE TCI, 2016), which is about $80\times$ faster than brute-force computation. This particular denoiser can be expressed as the proximal map of a convex regularizer and, as a consequence, we can guarantee convergence for linearized plug-and-play ADMM. We demonstrate the effectiveness of our proposals for super-resolution and single-photon imaging.
\end{abstract}

% ----------------- Keywords -------------------------

\section{Introduction}

Following the significant progress in image denoising, researchers have experimented with the idea of using modern denoisers for image restoration problems such as deconvolution, deblurring, tomography, and compressed sensing \cite{NCB2004,TMB2015,DKE2012,MMB2015}. Generally speaking, these are based on iterative methods, where each iteration involves the inversion of the forward model followed by the application of a powerful denoiser. Despite their empirical success, it has generally been challenging to furnish guarantees on convergence and optimality. The technical hurdle in this regard stems from the fact that state-of-the-art denoisers such as NLM \cite{BCM2005} and BM3D \cite{DFKE2007} are derived from a filtering and not an optimization perspective. To be precise, it is not known if they can be expressed as the proximal map \cite{boyd} of some regularizer $g(\x)$, i.e., whether 
the minimizer of 
\begin{equation}
\label{proxg}
g(\x) +  \beta \Vert \x - \tilde{\x} \Vert^2 \qquad \quad (\beta > 0)
 \end{equation} 
corresponds to the output obtained by denoising $\tilde{\x}$.

It was recently shown in \cite{SVWBDSB2016} that the existence of such a regularizer can be guaranteed for a doubly stochastic variant of NLM. In particular, it was shown that if $g(\x)$ is the regularizer in question and $f(\x)$ is data fidelity term associated with the forward model, then the ADMM-based solution \cite{boyd} of
\begin{equation}
\label{LSmodel}
 \underset{\x}{\text{min}} \ \ f(\x) +  \ \lambda g(\x) \quad \qquad (\lambda > 0)
\end{equation} 
provides a framework where the inversion and regularization steps can be decoupled, and the latter amounts to denoising using doubly stochastic NLM (see Section \ref{LPnP} for a detailed description). In fact, the idea of replacing the regularization step in ADMM with an arbitrary denoiser was proposed earlier in \cite{VBW2013} under the name ``plug-and-play'' ADMM (PnP-ADMM), albeit without any convergence guarantees. Later, it was also shown in \cite{CWE2017} that the iterates of PnP-ADMM are guaranteed to converge to a fixed point if the denoiser satisfies a certain boundedness property. However, the question of optimality of the fixed point was not resolved in \cite{CWE2017}. Moreover, it is not known if any of the existing denoisers satisfy this property.

%\subsection{Related Work and Contributions}

The present work has two-fold contributions. First, we propose to ``linearize'' the data term $f(\x)$ for the proximal update
\begin{equation}
\label{proxf}
 \underset{\x}{\text{min}} \  \ f(\x) +  \beta \Vert \x - \tilde{\x} \Vert^2
 \end{equation} 
used in PnP-ADMM \cite{SVWBDSB2016}. Linearization is used in several algorithms such as ISTA \cite{DDD2004}, FISTA \cite{BT2009}, and ADMM \cite{OCLP2015}. However, to the best of our knowledge, this has not been exploited for PnP-ADMM. Linearization allows us to perform the inversion updates at lower cost (without sacrificing convergence guarantees) for applications where \eqref{proxf} has to be computed in an iterative fashion. We note that techniques to cut down the inversion cost have been proposed in \cite{CWE2017,O2017}. It was shown in \cite{O2017} that the inversion can be performed efficiently using primal-dual splitting, and fast inversion techniques for specific restoration problems were proposed in \cite{CWE2017}. Our proposal requires $f(\x)$ to be differentiable. Moreover, we can guarantee convergence if $\nabla \! f(\x)$ is Lipschitz continuous. This is indeed the case for a wide range of problems including linear inverse problems \cite{DDD2004}. Importantly, we can incorporate hard constraints in the  optimization framework, which is difficult to do in PnP-ADMM \cite{SVWBDSB2016}. We note that even though $\nabla\! f(\x)$ fails to be Lipschitz for single-photon imaging \cite{F2011}, linearized ADMM is found to be stable and yields good reconstructions that are comparable with \cite{CWE2017} (see Section \ref{SPI}).

In a different direction, we also develop a fast low-complexity algorithm for doubly stochastic NLM (DSG-NLM). 
In \cite{SVWBDSB2016}, the authors conceived DSG-NLM as a matrix multiplication $\W \x$, where $\x$ is the vectorized input image and $\W$ is a weight matrix computed from patches. However, $\W$ is derived from the weight matrix of original NLM \cite{BCM2005} in three steps (see Section \ref{DNLM}). As a result, it was not apparent in \cite{SVWBDSB2016} that $\W \x$ can be computed using filtering (aggregation), without storing $\W$. We show that this can indeed be done using aggregation, however the aggregation needs to be performed not once but thrice. We also use an existing algorithm \cite{Darbon2008} for efficiently computing the patch distances. On the overall, we are able to accelerate the implementation of DSG-NLM by about $80\times$, for which PnP-ADMM (and the proposed linearization) comes with convergence and optimality guarantees. 

%\subsection{Organization}

In Section \ref{LPnP}, we propose linearized PnP-ADMM and discuss its convergence properties. 
A filtering perspective of DSG-NLM is presented in Section \ref{DNLM}, which is used to develop a fast algorithm.
In Section \ref{Exp}, we present results for super-resolution and single-photon imaging. 
We conclude with a summary of the results in Section \ref{Conc}.

\section{Linearized Plug-and-Play ADMM}
\label{LPnP}

The plug-and-play framework was originally proposed in \cite{VBW2013}. Following this original work, its algorithmic properties and applications have been studied in a series of papers \cite{SVWBDSB2016,BRE2016,RGE2016,CWE2017,wang2017,O2017}. The core idea is based on alternating direction method of multipliers (ADMM), which is tailored for solving composite optimization problems of the form in \eqref{LSmodel}, e.g., see \cite{boyd,ABF2010,OCLP2015,TBF2016}. In this work, we will consider the optimization model
\begin{equation}
\label{constrained}
 \underset{\x \in \mathbb{R}^n}{\text{min}} \ \ f(\x) +  \ \lambda g(\x) \quad \text{s.t.} \quad \x \in C,
\end{equation} 
where the data term $f(\x)$ forces consistency w.r.t. the measurements, the regularizer $g(\x)$ enforces some prior, and $C \subset \mathbb{R}^n$ is  closed and convex. %This is more general model than \eqref{LSmodel} and reduces to it when $C=\mathbb{R}^n$. 
Constraints of interest include $C=\mathbb{R}^n_+$ (non-negativity constraint) and $C=[0,1]^n$ (box constraint) \cite{ABF2010}. 
We next split the variable in \eqref{constrained} to obtain the following equivalent problem:
\begin{equation}
\label{split}
\begin{aligned}
\underset{\x,\v \in \mathbb{R}^n}{\text{argmin}} \ \  f(\x)  + \ \lambda g(\v) \quad \text{s.t.} \quad \x =  \v \ \text{ and } \ \x \in C.
\end{aligned}
\end{equation} 
The important point here is that though $\x$ is constrained, $\v$ (used in the denoising step) is a free variable which is indirectly constrained by the relation $\x=\v$. The augmented Lagrangian for \eqref{split} is given by 
\begin{equation*}
\L(\x,\v,\u) = f(\x) + \lambda g(\v) + \frac{\rho}{2} \Vert \x - \v + \u \Vert^2 - \frac{\rho}{2} \Vert \u \Vert^2,
\end{equation*}
where $\u$ is the (scaled) Lagrange multiplier and  $\rho > 0$ is the penalty parameter \cite{boyd}. The ADMM solution of \eqref{LSmodel} consists of the minimization of $\L(\x, \v^k,\u^k)$ over $\x \in C$:
\begin{equation}
\label{inv}
\x^{k+1} = \underset{\x \in C}{\text{argmin}} \  \ f(\x) + \frac{\rho}{2}\Vert \x - \tilde{\x}^k \Vert^2,
\end{equation}
the minimization of $\L(\x^{k+1}, \v,\u^k) $ over $\v \in \mathbb{R}^n$:
\begin{equation}
\label{reg}
\v^{k+1} = \underset{\v \in \mathbb{R}^n}{\text{argmin}}  \  \  \frac{\rho}{2 \lambda}\Vert \v - \tilde{\v}^k \Vert^2 + g(\v) ,
\end{equation}
where $\tilde{\x}^k = \v^k - \u^k$ and $\tilde{\v}^k = \x^{k+1} + \u^k$, and the dual update
\begin{equation*}
\u^{k+1} = \u^{k} + (\x^{k+1}-\v^{k+1}).
\end{equation*}
Notice that \eqref{reg} is the maximum-a-posteriori estimator corresponding to the observation $\tilde{\v}^k = \v + \n$ and prior $p(\v) \propto \exp(-g(\v))$, where $\n$ is iid Gaussian with variance $\sigma^2=\lambda/\rho$ \cite{hunt1977}. Different choices of $g(\v)$ lead to different denoising methods, such as wavelet and total-variation denoising \cite{ABF2010}. The key idea in \cite{VBW2013} was to substitute step \eqref{reg} with
\begin{equation}
\label{denoise}
\v^{k+1} = \D_{\sigma}(\tilde{\v}^k),
\end{equation}
where $\D_{\sigma}$ is some powerful denoiser such as NLM or BM3D (optimized to operate at noise level $\sigma$) that do not arise from any known regularizer. It was empirically demonstrated in \cite{VBW2013,SVWBDSB2016,BRE2016,RGE2016,CWE2017,wang2017} that this ad-hoc modification of the ADMM algorithm produces promising results for image restoration. Using the theory of proximal maps \cite{moreau1965}, it was later demonstrated in \cite{SVWBDSB2016} that one can associate a convex regularizer with NLM provided its weight matrix is modified to be symmetric and doubly stochastic, with eigenvalues in $[0,1]$ (discussed in detail in Section \ref{DNLM}). Moreover, it was shown that convergence is guaranteed in this case. 

In general, the optimization in \eqref{inv} has to be performed iteratively, even when $C=\mathbb{R}^n$. That is, along with the outer iterations, we need to perform inner iterations for \eqref{inv}. This is particularly applicable for the applications in Section \ref{Exp}. Ideally, we would like to replace  \eqref{inv} by a simple low-complexity operation.
%This is true even for linear inverse problems where $f(\x) = \lVert \A \x - \b \rVert^2$, $\A$ is some linear degradation, and $\b$ is the observed image. In this case, step \eqref{inv} requires us to solve a large linear system with coefficients $\A^\top\!\A + \rho \mathbf{I}$, which is a computational bottleneck. This can be done efficiently using FFT, but only for specific linear degradations \cite{ABF2010}. 
This can be achieved using a variant of ADMM called linearized ADMM \cite{OCLP2015}. The idea is to replace $f(\x)$ by its linear 
approximation around $\x^k$:
\begin{equation}
\label{linapp}
f(\x^k) +  \nabla\! f(\x^k)^\top \!(\x-\x^k) +  \frac{\alpha}{2}\Vert \x - \x^k \Vert^2,
\end{equation}
where we also append a quadratic term ($\alpha > 0$) for technical reasons to be discussed shortly. Substituting \eqref{linapp} in \eqref{inv}, we obtain
\begin{equation*}
\x^{k+1} = \underset{\x \in C}{\text{argmin}} \  \ \Vert \x - \bar{\x}^k \Vert^2 = \Pi_C(\bar{\x}^k),
\end{equation*} 
that is, $\x^{k+1}$ is the orthogonal projection of $\bar{\x}^k$ onto $C$, where $$\bar{\x}^k = (\alpha + \rho)^{-1} \big(\alpha  \x^k + \rho \tilde{\x}^k - \nabla \! f(\x^k)\big).$$ We note that $\Pi_C$ can be easily computed when $C$ is $\mathbb{R}^n_+$ or $[0,1]^n$. The assumption with linearization is that $f(\x)$ is continuously differentiable and $\nabla \! f(\x)$ is Lipschitz (and easily computable). This is true for a wide range of image restoration problems including linear inverse problems. The plug-and-play algorithm using linearized ADMM is summarized below. 

\SetKwFor{Loop}{loop}{}{end}
\begin{algorithm}
\label{ADMM}
\caption{Linearized Plug-and-Play ADMM.}
	\DontPrintSemicolon
	\KwIn{$\x, \v, \u$ and $\alpha, \lambda, \rho$.}
	Set $\mu =1/(\alpha + \rho)$ and $\sigma = \sqrt{\lambda/\rho}$.\;
	\While{not converged}
	{
$\x \longleftarrow \mu \big(\alpha \x + \rho (\v - \u) - \nabla \! f(\x)\big)$ \;
$\x \longleftarrow \Pi_C(\x)$ \;
$\v \longleftarrow \D_{\sigma}(\x + \u )$ \;
$\u \longleftarrow  \u + (\x-\v) $ \;
	}
\end{algorithm}
If $\D_{\sigma}$ is DSG-NLM, then Algorithm \ref{ADMM} is guaranteed to converge under some mild assumption on $f(\x)$. Indeed, as shown in \cite{SVWBDSB2016}, we can associate a closed, proper, and convex regularizer $g(\x)$ with $\D_{\sigma}$ in this case, i.e., \eqref{denoise} corresponds to the ADMM update \eqref{reg}. Algorithm \ref{ADMM} is then simply an application of linearized ADMM to problem \eqref{constrained}, and its convergence follows from existing  results \cite{OCLP2015}. Specifically, if $f(\x)$ is closed, proper, and convex (and satisfies some mild assumptions \cite{OCLP2015}), and $\alpha$ is larger than the Lipschitz constant of $\nabla \! f(\x)$, then $f(\x^k)+\lambda g(\x^k)$ converges to the optimum of \eqref{constrained}.

\section{Fast DSG-NLM}
\label{DNLM}

We now propose a fast algorithm for DSG-NLM \cite{SVWBDSB2016}. This denoiser is derived from NLM \cite{BCM2005} which uses patches and nonlocal aggregation for denoising. We will follow the notation in \cite{SVWBDSB2016} for easy comparison. In particular, we will use $\tilde{\v}$ and $\hat{\v}$ to denote the input to the denoiser and the output, both defined on a finite domain $S \subset \mathbb{Z}^2$. Moreover, we will use $\P_s$ (vector of length $N_p^2$) to denote a patch of size $N_p \times N_p$ centered at $s \in S$. In NLM, $\hat{\v}=\D_{\sigma}(\tilde{\v})$ is given by
\begin{equation}
\label{NLM}
\hat{\v}_s = \frac{\sum_{r \in \Omega_s} k_{s,r}   \tilde{\v}_r}{\sum_{r \in \Omega_s} k_{s,r} },  \quad  k_{s,r} = \exp \left( - \frac{\lVert \P_s - \P_r \rVert^2}{2 N_{p}^{2} \sigma^2}  \right),
\end{equation}
where $\Omega_s$ is a search window of size $(2N_s+1)^2$ centered at $s$ \cite{SVWBDSB2016}. 

Notice that we can express \eqref{NLM} as a linear transform $\hat{\v} = \K \tilde{\v}$, where $\K$ (derived from $\tilde{\v}$) is row stochastic, i.e., the sum of entries in each row is one. However, due to the division in \eqref{NLM}, $\K$ is not symmetric in general. It was shown in \cite[Section IV]{SVWBDSB2016} that, in three simple steps, we can transform $\K$ into a symmetric, doubly stochastic matrix $\W$ whose eigenvalues are in $[0,1]$. DSG-NLM is simply the transform  $\W \tilde{\v}$. For completeness, we recall these steps, but slightly differently from \cite{SVWBDSB2016}. In the first step, we set $\W=[w_{s,r}]$, where 
\begin{equation}
\label{S1}
w_{s,r} \leftarrow  \Lambda_{s,r}  k_{s,r}, \qquad  \Lambda_{s,r} = \Lambda\left(\frac{s-r}{N_s+1}\right),
\end{equation}
and $\Lambda(s) = (1-\lvert s_1\lvert)(1-\lvert s_2\lvert)$ is the separable hat function. We refer the reader to \cite{SVWBDSB2016} for the technical reason behind this and subsequent steps. The next step involves row and column normalization:
\begin{equation}
\label{S2}
w_{s,r} \leftarrow w_{s,r} \left(\sum_{r \in \Omega_s} w_{s,r} \right)^{-\frac{1}{2}} \left( \sum_{s \in \Omega_r} w_{s,r}\right)^{-\frac{1}{2}},
\end{equation}
We next compute the row sums of the weight matrix (at this point) and set the maximum to $1/\alpha$. The final step is given by
\begin{equation}
\label{S3}
w_{s,r} \leftarrow \alpha w_{s,r} \quad \text{and}  \quad w_{s,s} \leftarrow 1 + w_{s,s} - \sum_{r \in \Omega_s} w_{s,r} .
\end{equation}
It is proved in \cite[Theorem IV.1]{SVWBDSB2016} that the final matrix $\W$ is symmetric and doubly stochastic, and its eigenvalues are in $[0,1]$. 
As for the implementation, it is impractical to store the large matrix $\W$ and apply it on $\tilde{\v}$ as a matrix-vector multiplication. The present observation is that we can efficiently compute $\W   \tilde{\v}$ without having to store $\W$. However, unlike NLM, we need to loop over the pixels three times instead of just once. The complete procedure is provided in Algorithm \ref{fastNLM}, where images $\g$ and $\h$ are of the size of $\tilde{\v}$. The first aggregation corresponds to \eqref{S1}. In the second aggregation, we compute \eqref{S2} and the maximum of the row sums. In the final aggregation, we combine step \eqref{S3} and the application of $\W$ on $\tilde{\v}$.

\IncMargin{2mm}
\begin{algorithm}
\KwIn{$\tilde{\v}$.}
\KwOut{$\hat{\v}=\W \tilde{\v}$.}
Initialize $\hat{\v}, \g$, and $\h$ with zeros; also set $m = 0$\;
\For{$s \in S$}{ \label{start1}
    
     \lFor{$r \in \Omega_s$}{
	$\g_s \leftarrow \g_s +  \Lambda_{s,r} k_{s,r} $
}
} \label{end1}
\For{$s \in S$}{
    $\delta \leftarrow  0$\;
     \lFor{$r \in \Omega_s$}{
	$\delta \leftarrow \delta + (\g_s  \g_r)^{-\frac{1}{2}}\Lambda_{s,r}  k_{s,r} $} 
$m \leftarrow  \text{max}(m, \delta)$ \;
}
\For{$s \in S$}{
     \For{$r \in \Omega_s$}{
	$w \leftarrow  m^{-1}  (\g_s  \g_r)^{-\frac{1}{2}}\Lambda_{s,r}  k_{s,r} $\;
	$\hat{\v}_s \leftarrow \hat{\v}_s + w \ \tilde{\v}_r$\;
	$\h_s \leftarrow \h_s + w $\;
}
$\hat{\v}_s \leftarrow \hat{\v}_s + (1 - \h_s) \tilde{\v}_s$.
}
\caption{DSG-NLM Filtering.}
\label{fastNLM}
\end{algorithm}
\DecMargin{2mm}

Similar to NLM, it is evident that the complexity of Algorithm \ref{fastNLM} is $O(N_s^2 N_p^2)$ per pixel. However, similar to NLM, we can use the fast algorithm in \cite{Darbon2008} for computing patch distances, whose complexity does not depend on the patch size $N_p$. This significantly brings down the complexity to $O(N_s^2)$. In summary, we can compute DSG-NLM via filtering and that too at a reduced complexity. 

\section{Experiments}
\label{Exp}

We validate the performance of our restoration algorithm using super-resolution \cite{PPK2003} and single photon imaging \cite{F2011}. 
The reason behind this choice is that the data fidelity term  is quadratic in super-resolution and non-quadratic (though convex and differentiable) in single photon imaging. Assuming (without loss of generality) that the intensity of the ground-truth is in $[0,1]$, we set the constraint set $C$ in \eqref{constrained} to be $C=[0,1]^n$ 
For both applications, we have used BM3D, NLM and DSG-NLM for denoising. 
Moreover, similar to \cite{SVWBDSB2016}, we have tried both the adaptive (A-DSG-NLM) and fixed (F-DSG-NLM) variants of DSG-NLM.
In the former, the weight $\W$ is adapted in each iteration as discussed previously.
In F-DSG-NLM, we stop adapting $\W$ after certain number of iterations (typically $15$), which is used for rest of the iterations.
The point to note is that convergence is guaranteed for F-DSG-NLM but not for A-DSG-NLM, since the output is not a linear function of the input in the latter  \cite{SVWBDSB2016}.

In Table  \ref{timingNLM}, we compare the timings of DSG-NLM \cite{SVWBDSB2016} and Algorithm \ref{fastNLM} for a $256 \times 256$ image. 
The simulation was done using Matlab on a quad-core $3.4$ GHz machine with $16$ GB memory.
We used standard patch and window sizes \cite{BCM2005}. 
The brute-force implementation of nonlocal denoising is known to be prohibitively slow \cite{Darbon2008}.
In Table  \ref{timingNLM}, notice that the brute-force implementation of DSG-NLM takes minutes, while Algorithm \ref{fastNLM} takes just few seconds.
The speedup is almost $100\times$ for large patch sizes. 
This is because the complexity of our fast algorithm does not scale with the patch size.
 
\subsection{Single image super-resolution}

The forward model for single image super-resolution is given by
\begin{equation*}
\label{model}
\y = \A\x + \n,
\end{equation*}
where $\x$ is the high-resolution image (ground truth), $\A$ is a linear operator, and $\n$ is iid Gaussian noise \cite{PPK2003}. Specifically, $\A\x$ is the low-pass filtering of $\x$ (using blur $\boldsymbol{h}$) followed by downsampling (by factor $k$). The adjoint $\A^\top\!$ corresponds to upsampling (by factor $k$) and low-pass filtering with $\boldsymbol{h}$ (if $\boldsymbol{h}$ is symmetric) \cite{ABF2010}. The problem is to recover $\x$ from the low-resolution image $\y$. As is well-known \cite{hunt1977}, the data fidelity term corresponding to the negative log-likelihood of $\y$ given $\x$ is
\begin{equation}
\label{dataterm}
f(\x) = \frac{1}{2} \Vert \y - \A\x \Vert^2.
\end{equation}
The gradient and Hessian of  \eqref{dataterm} are $\A^\top\!(\y - \A\x)$ and  $\A^\top\!\A$. The Lipschitz constant of $\nabla \!f(\x)$ is simply the largest eigenvalue of $\A^\top\!\A$  \cite{ABF2010}. If we set $\alpha > \lambda_{\text{max}}(\A^\top\!\A)$ and use A-DSG-NLM for denoising, then convergence is guaranteed for Algorithm \ref{ADMM}.

\begin{figure}[H]
\centering
\subfloat[]{\includegraphics[width=0.325\linewidth]{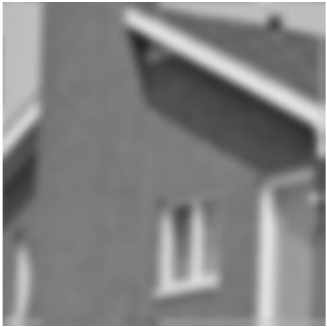}} \hspace{0.1mm}
\subfloat[]{\includegraphics[width=0.325\linewidth]{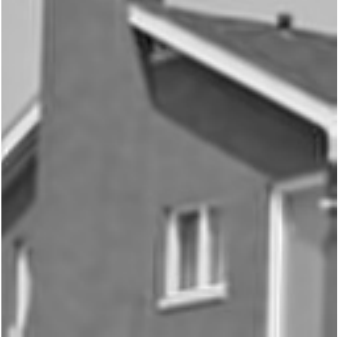}} \hspace{0.1mm}
\subfloat[]{\includegraphics[width=0.325\linewidth]{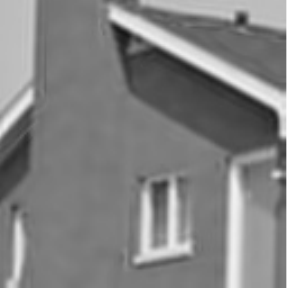}} \hspace{0.1mm}
\caption{Super-resolution results for \textit{House} using $k=2$ (periodic boundary). The standard deviations of Gaussian blur and noise are $1.5$ and $2/255$. (a) Cubic interpolation, \textbf{PSNR} = $27.09$ dB; (b) Chan \cite{CWE2017}, \textbf{PSNR} = $32.72$ dB; and (c) Proposed, \textbf{PSNR} = $32.61$ dB.}
\label{sr}
\end{figure}

\begin{table}
\scalebox{0.95}{
\begin{tabular}{l*{5}{c}r}
Patch Size ($N_p$)             & $11 \times 11$ & $17 \times 17$ & $23 \times 23$ & $29 \times 29$ \\
\hline
\vspace*{-0.3cm}\\
Proposed   & $3.90$  & $4.01$ & $4.12$ & $4.25$ \\
Brute-force    & $299.54$  & $326.03$ & $366.65$ & $439.19$ \\
\end{tabular}}
\caption{Timings (sec) of the brute-force and fast implementations of DSG-NLM for different patch size and fixed window size $N_s=21$.}
\label{timingNLM}
\end{table}

\begin{table}[H]
\scalebox{0.85}{
\begin{tabular}{l*{6}{c}r}
Error              & BM3D & NLM &  A-DSG-NLM & F-DSG-NLM \\
\hline
\vspace*{-0.3cm}
\\
Primal & $3.10 \times 10^{-3}$  & $7.30 \times 10^{-4}$  & $1.36 \times 10^{-5}$ & $2.44 \times 10^{-8}$  \\
Dual  & $1.01 \times 10^{-1}$  & $4.73 \times 10^{-2}$ & $1.42 \times 10^{-4}$ & $2.81 \times 10^{-6}$  \\
PSNR                  & $32.43$ & $31.10$ & $32.56$ & $32.61$  \\
\end{tabular}}
\caption{Primal and dual residuals and corresponding PSNRs (after 250 iterations) for image super-resolution (periodic boundary).}
\label{srtable}
\end{table}

\begin{figure}
\centering
\includegraphics[width=0.90\linewidth]{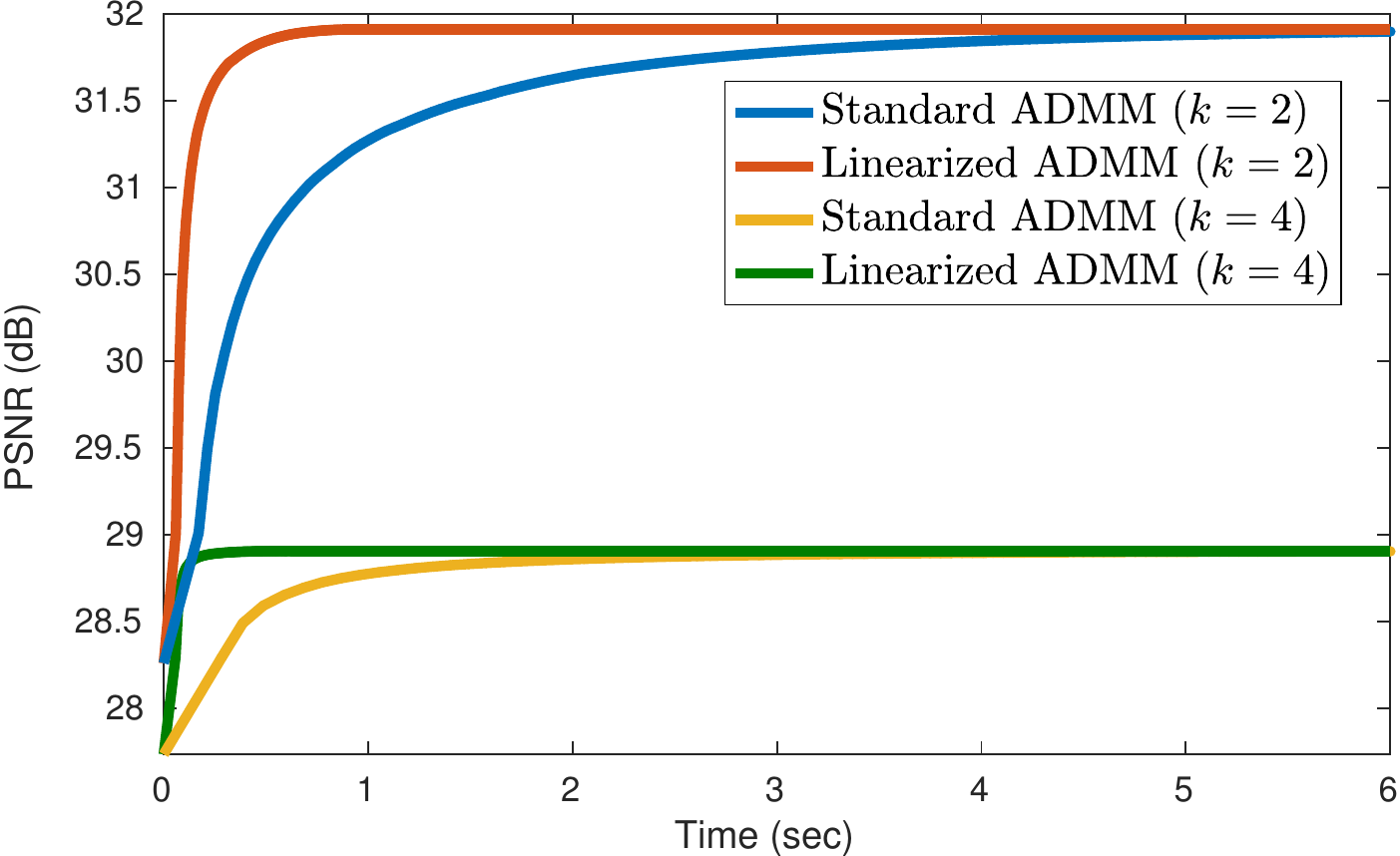}
\caption{PSNR evolution for super-resolution (symmetric boundary) with downsampling factor $k$ using standard and linearized ADMM.}
\label{psnr_evolution}
\end{figure}

We now present some results for image super-resolution where $\h$ (blur) is Gaussian. A typical result is reported in Figure \ref{sr}, where we have also compared with the plug-and-play algorithm in \cite{CWE2017}. For both methods, we have used F-DSG-NLM as the denoiser. In \cite{CWE2017}, the inversion step \eqref{inv} is computed using fast polyphase decompositions and $\rho$ is adapted at each iteration. Notice that the reconstructions are comparable, both visually and in terms of PSNR. We note that, unlike Algorithm \ref{ADMM}, the fast inversion in \cite{CWE2017} works just with periodic convolutions. Following this observation, we perform a super-resolution experiment for the setup in Figure \ref{sr}, but using symmetric boundary condition (for the blur). In this case, the inversion step \eqref{inv} requires the solution a linear system with coefficients $\A^\top\!\A + \rho \mathbf{I}$. This can be done iteratively using conjugate-gradients, which is however computation intensive. The evolution of PSNR with time for standard and linearized plug-and-play ADMM are compared in Figure \ref{psnr_evolution}. Notice that the PSNR peaks very fast in linearized ADMM. A possible explanation is that though we use an inexact update for  \eqref{inv}, this is partly compensated by the regularization (smoothing) step in each iteration.

\subsection{Single photon imaging}
\label{SPI}

\begin{figure}
\centering
\subfloat[]{\includegraphics[width=0.325\linewidth]{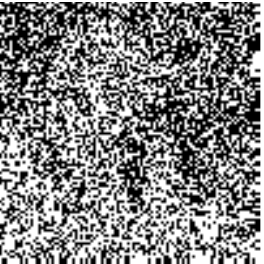}} \hspace{0.1mm}
\subfloat[]{\includegraphics[width=0.325\linewidth]{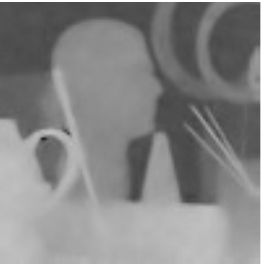}} \hspace{0.1mm}
\subfloat[]{\includegraphics[width=0.325\linewidth]{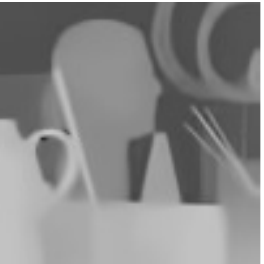}} \hspace{0.1mm}
\caption{Single photon imaging with $K =16$. (a) Binary measurements; (b) Chan \cite{CWE2017} with F-DSG-NLM, \textbf{PSNR} = $32.60$ dB; and (c) Proposed with F-DSG-NLM, \textbf{PSNR} = $33.24$ dB.}
\label{photon}
\end{figure}

We next consider single photon imaging using quanta image sensors (QIS) \cite{F2011}. 
For an input image $\x \in \mathbb{R}^n$, the QIS consists of $nK$ photon detectors, which are uniformly distributed to read the incoming light ($K$ is the oversampling factor). A simplified model of the amount of photon $\s$ arriving at QIS pixels is given by $\s = \eta \G \x $, where $\eta$ is the sensor gain, and $\G = (1/K) (\mathbf{I}_{n \times n} \otimes \mathbf{1}_{K \times 1})$. 
More precisely, the photon count follows a Poisson distribution with parameter $\s$. The final output $\y$ is a binary image obtained by thresholding the photon count (see \cite{CWE2017} for a detailed description). In brief, the negative log-likelihood of observing $\y$ given $\x$ is given by
\begin{equation}
\label{obj}
f(\x) = \sum_{i = 1}^{n} h(x_i),
\end{equation}
where 
\begin{equation*}
h(t) = -K_i^0 \log\big(e^{-\frac{\eta t}{K}}\big) -K_i^1 \log\big(1-e^{-\frac{\eta t}{K}}\big),
\end{equation*}
and $K_i^0, K_i^1$ are known design parameters. Since \eqref{obj} is separable, its gradient  is 
$\nabla \! f(\x) = \left(h'(x_1),\ldots,h'(x_n)\right)$, where
\begin{equation*}
\label{photongrad}
h'(t) = \frac{\eta}{K} \left(K_i^0 - K_i^1 e^{-\frac{\eta t}{K}} \left(1-e^{-\frac{\eta t}{K}}\right)^{-1} \right).
\end{equation*}
It can be verified that $h''(t)$ is not bounded near the origin. Therefore, $\nabla \! f(\x)$ cannot be Lipschitz. However, this is only a sufficient condition for convergence, and Algorithm \ref{ADMM} in fact was found to be work stably for our experiments. As shown in Figure \ref{photon}, the reconstruction obtained using Algorithm \ref{ADMM} is comparable to that obtained using \cite{CWE2017}.  

\begin{table}[h]
\scalebox{0.85}{
\begin{tabular}{l*{6}{c}r}
Error               & BM3D & NLM &  A-DSG-NLM & F-DSG-NLM \\
\hline
\vspace*{-0.3cm}
\\
Primal & $6.39 \times 10^{-4}$  & $1.22 \times 10^{-10}$  & $1.72 \times 10^{-13}$ & $1.83 \times 10^{-15}$  \\
Dual  & $1.28 \times 10^{-2}$  & $1.36 \times 10^{-8}$ & $1.32 \times 10^{-11}$ & $1.35 \times 10^{-12}$  \\
PSNR                  & $32.55$ & $30.98$ & $32.60$ & $33.24$  \\
\end{tabular}}
\caption{Same as in Table \ref{srtable} but for single photon imaging.}
\label{photontable}
\end{table}

For both experiments, we computed the primal and dual residuals \cite{boyd} and the PSNR for different denoisers. The results are reported in Tables \ref{srtable} and \ref{photontable}. We observe that the residuals for BM3D, NLM, and to some extent A-DSG-NLM, do not seem to converge fully. 
On the other hand, the residual for F-DSG-NLM is several orders smaller than NLM and BM3D. For single photon imaging, the residual for F-DSG-NLM is $4$ orders smaller than NLM and $10$ orders smaller than BM3D. Interestingly, the PSNR for F-DSG-NLM is better than that of BM3D or NLM (a similar observation was reported in \cite{SVWBDSB2016}).

\section{Conclusion}
\label{Conc}

We proposed a plug-and-play algorithm for image restoration, where both the inversion and denoising steps can be computed efficiently. 
The proposed algorithm comes with convergence guarantees for linear inverse problems.
For super-resolution and single photon imaging, our algorithm was shown to yield reconstructions that are comparable with existing plug-and-play algorithms. Though this was not investigated due to space constraints, we can use accelerated variants of linearized ADMM to further speed up the reconstruction  \cite{OCLP2015}. Another interesting possibility is to use improved variants of nonlocal means \cite{wu2013,ghosh2017} for the denoising step. 

\clearpage

% ----------------- References -----------------------
\bibliographystyle{IEEEtran}
\bibliography{citations}
% -----------------------------------------------------

\end{document}